\pdfoutput=1

\documentclass[11pt]{article}

\usepackage[]{ACL2023}

\usepackage{times}
\usepackage{latexsym}
\usepackage[T1]{fontenc}
\usepackage[utf8]{inputenc}
\usepackage{microtype}
\usepackage{inconsolata}
\usepackage{amsmath}
\usepackage{nicefrac}
\usepackage{tabularx}
\usepackage{amssymb}
\usepackage{multirow}
\usepackage{booktabs}
\usepackage{graphicx}

\title{Emergent Linear Representations in World Models of Self-Supervised Sequence Models}

\author{
Neel Nanda$^*$ \\
Independent
\And
Andrew Lee$^*$ \\
University of Michigan
\And
Martin Wattenberg \\
Harvard University
}

\begin{document}

\maketitle
\def\thefootnote{*}\footnotetext{Equal contribution. \texttt{neelnanda27@gmail.com}, \newline \texttt{ajyl@umich.edu}}\def\thefootnote{\arabic{footnote}}

\begin{abstract}

How do sequence models represent their decision-making process?
Prior work suggests that Othello-playing neural network learned nonlinear models of the board state \citep{li2023emergent}.
In this work, we provide evidence of a closely related \textit{linear} representation of the board.
In particular, we show that probing for ``my colour'' vs. ``opponent's colour'' may be a simple yet powerful way to interpret the model's internal state.
This precise understanding of the internal representations allows us to control the model's behaviour with simple vector arithmetic.
Linear representations enable significant interpretability progress, which we demonstrate with further exploration of how the world model is computed.\footnote{Code available at \url{https://github.com/ajyl/mech_int_othelloGPT}}

\end{abstract}

\section{Introduction}
\label{sec:intro}

How do sequence models represent their decision-making process?
Large language models are capable of unprecedented feats, yet largely remain inscrutable black boxes.
Yet evidence has accumulated that models act as feature extractors: identifying increasingly complex properties of the input and representing these in the internal activations \citep{geva-etal-2021-transformer, bau2020units, gurnee2023finding, 10.1162/coli_a_00422, burns2022dl, goh2021multimodal, elhage2022solu}.
A key first step for interpreting them is understanding how these features are represented.
\citet{mikolov2013linguistic} introduce the \textbf{linear representation hypothesis}: that features are represented linearly as directions in activation space.
This would be highly consequential if true, yet this remains controversial and without conclusive empirical justification.
In this work, we present novel evidence of linear representations, and show that this hypothesis has real predictive power.

We build on the work of \citet{li2023emergent}, who demonstrate the emergence of a \emph{world model} in sequence models.
Namely, the authors train OthelloGPT, an autoregressive transformer model, to predict legal moves in a game of Othello given a sequence of prior moves (Section~\ref{subsec:prelim:othelloGPT}).
They show that the model spontaneously learns to track the correct board state, recovered using \emph{non-linear} probes, despite never being told that the board exists.
They further show a causal relationship between the model's inner board state and its move predictions using model edits.
Namely, they show that the edited network plays moves that are legal in the edited board state even if illegal in the original board, and even if the edited board state is unreachable by legal play (i.e., out of distribution). 

\begin{figure}[t]
  \centering
    \includegraphics[clip,  width=0.98\columnwidth]{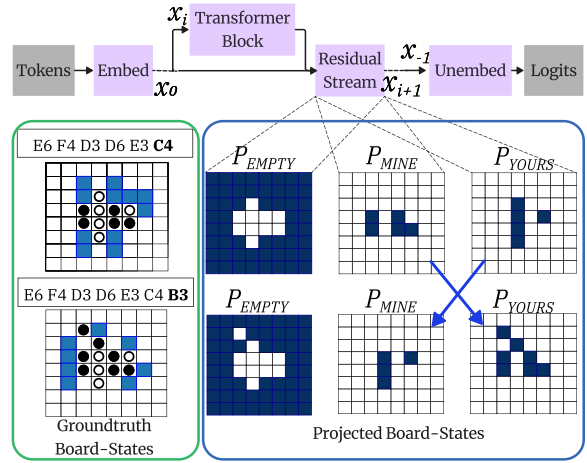}
  \caption{\label{fig:method_projection}
The emergent world models of OthelloGPT are linearly represented.
We find that the board states are encoded relative to the current player's colour (\textsc{Mine} vs. \textsc{Yours}) as opposed to absolute colours (\textsc{Black} vs. \textsc{White}).
}
\end{figure}

Critically, the original authors claim that OthelloGPT uses \emph{non-linear} representations to encode the board state, by achieving high accuracy with non-linear probes, but failing to do so using linear probes.
In our work, we demonstrate that a closely related world model is actually \emph{linearly} encoded.
Our key insight is that rather than encoding the \emph{colours} of the board (\textsc{Black}, \textsc{White}, \textsc{Empty}), the sequence model encodes the board \emph{relative} to the current player of each timestep (\textsc{Mine}, \textsc{Yours}, \textsc{Empty}).
In other words, for odd timesteps, the model considers \textsc{Black} tiles as \textsc{Mine} and \textsc{White} tiles as \textsc{Yours}, and vice versa for even timesteps (Section~\ref{sec:linear_reps}).
Using this insight, we demonstrate that a \emph{linear} projection can be learned with near perfect accuracy to derive the board state. 

We further demonstrate that we can steer the sequence model's predictions by simply conducting vectoral arithmetics using our linear vectors (Section~\ref{sec:intervene}).
Put differently, by pushing the model's activations in the directions of \textsc{Mine}, \textsc{Yours}, or \textsc{Empty}, we can alter the model's belief state of the board, and change its predictions accordingly.
Our intervention method is much simpler and interpretable than that of \citet{li2023emergent}, which rely on gradients to update the model's activations (Section~\ref{subsec:intervene_method}).
Our results confirm that our interpretation of each probe direction is correct, but also demonstrates that a mechanistic understanding of model representations can lead to better control.
Our results do not contradict that of \citet{li2023emergent}, but add to our understanding of emergent world models.

We provide additional interpretations of the sequence model using linear operations.
For example, we provide empirical evidence of how the model derives empty tiles of the board, and find additional linear representations, such as tiles being \textsc{Flipped} at each timestep.

Finally, we provide a short discussion of our thoughts.
How should we think of linear versus non-linear representations?
Perhaps most interestingly, why do linear representations emerge?

\section{Preliminaries}
\label{sec:prelim}

In this section we briefly describe Othello, OthelloGPT, and our notations.

\subsection{Othello}
\label{subsec:prelim:othello}

Othello is a two player game played on a 8x8 grid.
Players take turns playing black or white discs on the board, and the objective is to have the majority of one's coloured discs by the end of the game.

At each turn, when a tile is played, all of the opponent's discs that are enclosed in a horizontal, vertical, or diagonal row between two discs of the current player are flipped.
The game ends when there are no more valid moves for both players.

\subsection{OthelloGPT}
\label{subsec:prelim:othelloGPT}

OthelloGPT is a 8-layer GPT model \citep{radford2019language}, each layer consisting of 8 attention heads and a 512-dimensional hidden space.
We use the model weights provided by \citet{li2023emergent}, denoted there as the synthetic model.
The vocabulary space consists of 60 tokens, each one corresponding to a playable move on the board (e.g., A4).\footnote{The game always starts with 4 tiles in the center of the board already filled.}

The model is trained in an autoregressive manner, meaning for a given sequence of moves $m_{<t}$, the model must predict the next valid move $m_{t}$.

Note that no a priori knowledge of the game nor its rules are provided to the model.
Rather, the model is only given move sequences with a training objective to predict next valid moves.
Further note that these valid moves are uniformly chosen, and this training objective differs from that of models like AlphaZero \citep{alphazero}, which are trained to play strategic moves to win games.

\begin{table*}
\center
\begin{tabular}{ccccccccc}
\hline
    & $x^0$ & $x^1$ & $x^2$ & $x^3$ & $x^4$ & $x^5$ & $x^6$ & $x^7$  \\ 
\hline
Randomized    & 37 & 35.1 & 33.9 & 35.5 & 34.8 & 34.7 & 34.4 & 34.5 \\
Probabilistic & \multicolumn{8}{c}{61.8} \\
Linear \{\textsc{Black}, \textsc{White}, \textsc{Empty}\} & 62.2 & 74.8 & 74.9 & 75.0 & 75.0 & 74.9 & 74.8 & 74.4 \\
Non-Linear \{\textsc{Black}, \textsc{White}, \textsc{Empty}\} & 63.4 & 88.6 & 93.3 & 96.3 & 97.5 & 98.3 & 98.7 & 98.3 \\
\hline
Linear \{\textsc{Mine}, \textsc{Yours}, \textsc{Empty}\} & \textbf{90.9} & \textbf{94.8} & \textbf{97.2} & \textbf{98.3} & \textbf{99} & \textbf{99.4} & \textbf{99.6} & \textbf{99.5} \\
\hline
\end{tabular}
\caption{Probing accuracy for board states.
OthelloGPT linearly encodes the board state relative to the current player at each timestep (\textsc{Mine} vs. \textsc{Yours}, as opposed to colours \textsc{Black} or \textsc{White}.
}
\label{tab:linear_acc}
\end{table*}

\subsection{Notations}
\label{subsec:prelim:notations}

\paragraph{Transformers.}
Our transformer architecture \citep{NIPS2017_3f5ee243} consists of embedding and unembedding layers $Emb$ and $Unemb$ with a series of $L$ transformer layers in-between.
Each transformer layer $l$ consists of $H$ multi-head attentions and a multilayer perception (MLP) layer.

A forward pass in the model first embeds the input token at timestep $t$ using embedding layer $Emb$ into a high dimensional space $x_t^0 \in \mathbb{R}^{D}$.
We refer to $x_{t\in T}^0$ as the start of the \emph{residual stream}.
Then each attention head $Att_l^h, \forall h\in H$ and MLP block at layer $l$ add to the residual stream:

\begin{align*}
    x_t^{l\_mid} = x_t^l + \sum_{h\in H} Att_l^h(x_t^l)\\
    x_t^{l+1} = x_{t}^{l\_mid} + MLP_l(x_t^{l\_mid})
\end{align*}

Each attention head $Att^h_l$ computes value vectors by projecting the residual stream to a lower dimension using $Att^h_l.V$, linearly combines value vectors using $Att^h_l.A$, and projects back to the residual stream using $Att^h_l.O$:

\begin{align*}
    h(x) = (Attn^h_l.A \otimes Attn^h_l.O*Attn^h_l.V) * x
\end{align*}

A final prediction is made by applying $Unemb$ on $x^{L-1}$, followed by a softmax.

\paragraph{Probe Models.}
We notate linear and non-linear probes as $p^\lambda$ and $p^\nu$.
Our linear probes are simple linear projections from the residual stream: $p^\lambda(x_t^l)=\text{softmax}(Wx_t^l), W \in \mathbb{R}^{D\times 3}$.
The dimension $D\times3$ comes from doing a 3-way classification.\footnote{
In practice, because we are predicting the state of all 64 tiles, the shape of our probe is $D\times64\times3$.}
Non-linear probes are 2-layer MLP models: $p^\nu(x_t^l)=\text{softmax}(W_1 \text{ReLU}(W_2 x_t^l))$, $W_1 \in \mathbb{R}^{H\times 3}, W_2 \in \mathbb{R}^{D\times H}$.
\citet{li2023emergent} classify the colour at each tile (\textsc{Black}, \textsc{White}, \textsc{Empty}).
Our insight is to classify the colours \emph{relative} to the current turn's player (\textsc{Mine}, \textsc{Yours}, \textsc{Empty}).

\section{Linearly Encoded Board States}
\label{sec:linear_reps}

In this section we describe our experiments to find linear board state representations.

\subsection{Experiment Setup}
\label{subsec:linear_reps_experiments}

Rather than encoding the colour of each tile (\textsc{Black}, \textsc{White}, \textsc{Empty}), OthelloGPT encodes each tile \emph{relative} to the player of each timestep (\textsc{Mine}, \textsc{Yours}, \textsc{Empty}) --- for \emph{odd} timesteps, we consider \textsc{Black} to be \textsc{Mine} and \textsc{White} to be \textsc{Yours}, and vice versa for \emph{even} timesteps.

In order to learn the weights of our linear probe, we train on 3,500,000 game sequences.
We use a validation set of 512 games, and train until our validation loss converges according to a patience value of 10.
In practice, our linear probes converge after around 100,000 training samples.
We test our probes on a held out set of 1,000 games.

We train a different probe for each layer $l$.
Hyperparameters are provided in the Appendix.

\subsection{Results}
\label{subsec:linear_reps_results}

Table~\ref{tab:linear_acc} shows the accuracy for various probes.

We include four baselines.
The first is a linear probe trained on a randomly initialized GPT model.
We also include a probabilistic baseline, in which we always choose the most likely colour per tile at each timestep, according to a set of 60,000 games from training data.
The next two baselines are probe models used in \citet{li2023emergent}: a linear and non-linear probe trained to classify amongst \{\textsc{Black}, \textsc{White}, \textsc{Empty}\}.

Our linear probes achieve high accuracy by layer 4.
Unbeknownst previously, we show that the emerged board state is linearly encoded.

\begin{figure}[t]
  \centering
    \includegraphics[clip,  width=0.98\columnwidth]{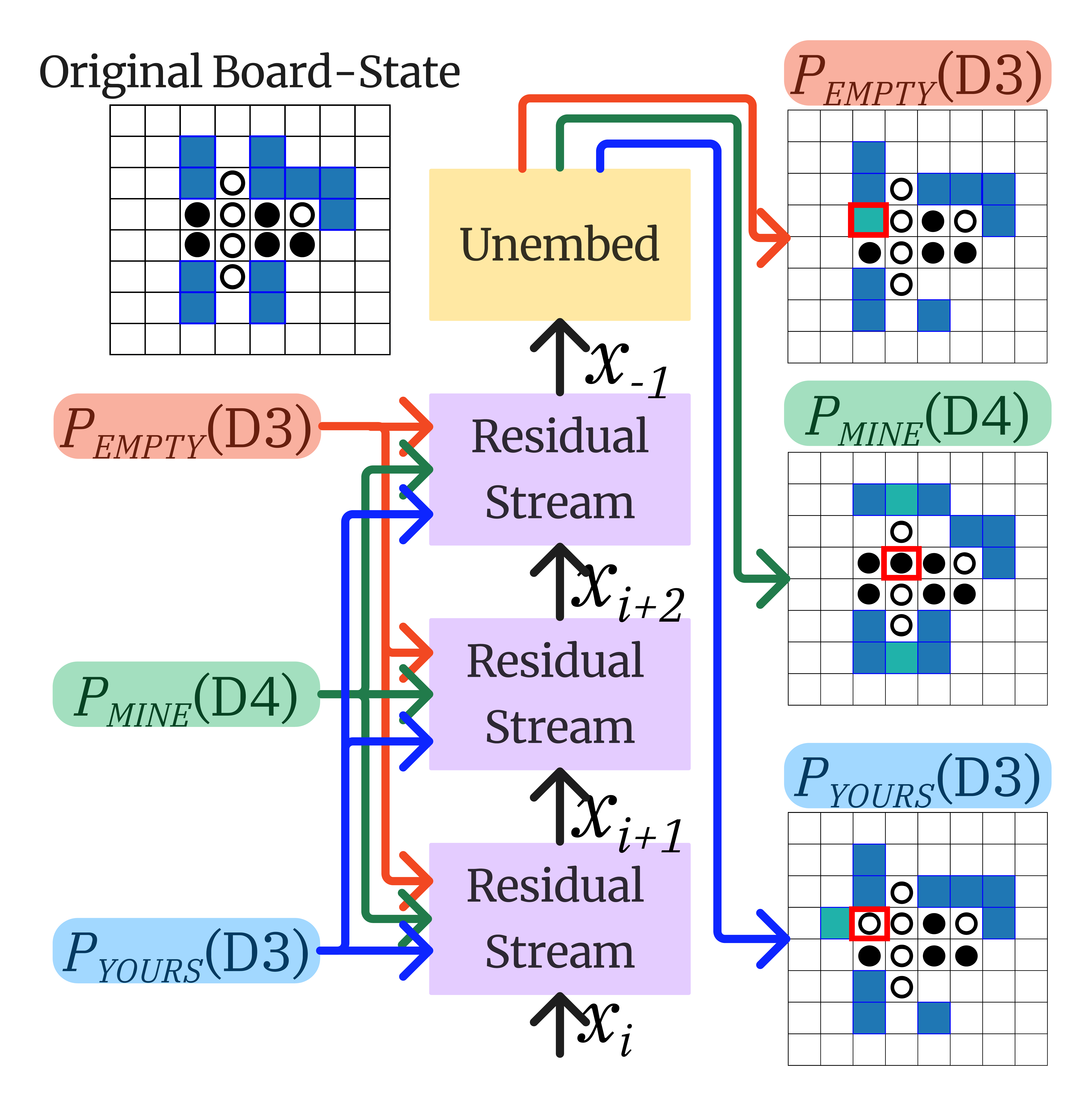}
  \caption{\label{fig:method_intervene}
Intervening methodology: we intervene by adding either \textsc{Empty}, \textsc{Mine}, or \textsc{Yours} directions into each layer of the residual stream.
Red squares in each board indicate the tiles that have been intervened, teal tiles indicate new legal moves post-intervention that the model predicts.
}
\end{figure}

\section{Intervening with Linear Directions}
\label{sec:intervene}

In this section we demonstrate how we intervene on OthelloGPT's board state using linear probes.

\subsection{Method}
\label{subsec:intervene_method}

An inherent issue with probing is that it is correlational, not causal \citep{belinkov-2022-probing}.
To validate that our probes have found a true world model, we confirm that the model uses the encoded board state for its predictions.

To verify this, we conduct the same intervention experiment as \citet{li2023emergent}.
Namely, given an input game sequence (and its corresponding board state $B$), we intervene to make the model believe in an altered board state $B'$.
We then observe whether the model's prediction reflects the made-believe board state $B'$ or the original board state $B$.

Our intervention approach is simple: we add our linear vectors to the residual stream of each layer:

\begin{align*}
    x' \leftarrow x + \alpha p^\lambda_d(x)
\end{align*}

where $d$ indicates a direction amongst \{\textsc{Mine}, \textsc{Yours}, \textsc{Empty}\} and $\alpha$ is a scaling factor.
In other words, to flip a tile from \textsc{Yours} to \textsc{Mine}, we simply push the residual stream at every layer in the \textsc{Mine} direction, or to ``erase'' a previously played tile, we push in the \textsc{Empty} direction.
\footnote{We experiment with intervening on different layers.
See Appendix for more details.}
\footnote{We use the TransformerLens library: \url{https://github.com/neelnanda-io/TransformerLens}.}

Note that this intervention is much simpler than that of \citet{li2023emergent}.
Namely, \citet{li2023emergent} edits the activation space ($x$) of OthelloGPT using several iterations of gradient descent from their non-linear probe.
Instead, we perform a single vector addition.

\subsection{Experiment Setup}
\label{subsec:intervene_experiment_setup}

For our intervention experiment, we adopt the same setup and metrics as \citet{li2023emergent}.
We use an evaluation benchmark consisting of 1,000 test cases.
Each test case consists of a partial game sequence ($B$) and a targeted board state $B'$.

We measure the efficacy of our intervention by treating the task as a multi-label classification problem.
Namely, we compare the top-$N$ predictions post-intervention against the groundtruth set of legal moves at state $B'$, where $N$ is the number of legal moves at $B'$.
We then compute error rate, or the number of false positives and false negatives.

\citet{li2023emergent} only considers the scenario of flipping the colour of a tile.
To also validate our \textsc{Empty} direction, we also experiment with ``erasing'' a previously played tile by making it empty.

\begin{table}
\center
\begin{tabular}{cc}
\toprule
Flipping colours    & Avg. \# Errors \\
\hline
Null Intervention Baseline  & 2.723 \\
Non-Linear Intervention     & 0.12 \\
Linear Probe Addition       & \textbf{0.10} \\
\midrule
Erasing         & Avg. \# Errors \\
\hline
Null Intervention           &   2.73 \\
Non-Linear Intervention     &   0.11 \\
Linear Probe Addition       &   \textbf{0.02} \\

\bottomrule
\end{tabular}
\caption{
Error rates from interventions.
}
\label{tab:intervention}
\end{table}

\subsection{Results}
\label{subsec:intervene_results}

Table~\ref{tab:intervention} shows the average error rates after our interventions.
Our interventions are equally effective as that of gradient-based editing, and confirms that our interpretation of each linear direction matches how the model uses such directions.

\section{Additional Linear Interpretations}
\label{section:other_linear_interps}

The linear representation hypothesis is of interest to the mechanistic interpretability community because it provides a foothold into understanding a system.
The internal state of the transformer, the residual stream, is the sum of the outputs of all previous components (heads, layers, embeddings and neurons) \citep{elhage2021mathematical}, so any linear function of the residual stream can be linearly decomposed into contributions from each component, allowing us to trace back where a computation is coming from.

In this section we leverage our newfound linear representation of board state to provide additional interpretations of OthelloGPT, as proof of concept of how discovering linear representations unlocks downstream interpretability applications.

\subsection{Interpreting Empty Tiles}
\label{subsec:empty_tiles}

Here we interpret how OthelloGPT derives the status of empty tiles.

\paragraph{The \textsc{Empty} Circuit.}
A key insight for \textsc{Empty} is that input tokens each correspond to a tile on the board (i.e., A4), and once played, the tile can only change colour but remains non-empty.

We view OthelloGPT as using attention heads to ``broadcast'' which moves have been played: given a move at timestep $t$, attention heads write this information into other residual streams.
This information (\textsc{Played}) can be represented as following.
First, each move $m$ (A4) is embedded: $Emb[m]$.
Then the model writes this information to other residual streams using linear projections $Att.V$ and $Att.O$ (Section~\ref{subsec:prelim:notations}):

\begin{align*}
    \textsc{Played}_h(m) = Emb[m] @ Att_h.V @ Att_h.O
\end{align*}

For each attention head in the first layer,\footnote{Knowing which moves were \textsc{Played} (i.e. show up in the input sequence), should not depend on any other computation, and thus we expect this information to be written by the attention heads in the first layer.} we compute the cosine similarity between \textsc{Played} and the $p^\lambda_{\textsc{empty}}$ direction:

\begin{align*}
    \max_{h \in H} \text{CosSim}(\textsc{Played}_h(m), p^\lambda_{\textsc{empty}}(m))
\end{align*}

Since the two terms encode \emph{opposite} information, we expect a high negative cosine similarity.

We observe an average similarity score of \textbf{-0.862} across all 60 squares,\footnote{The center 4 squares can never be empty.}, confirming that $p_{\textsc{Empty}}$ is encoding \textsc{Not Played}.
This tells us that $p_{\textsc{Empty}}$ is a linear function of the token embeddings.

This also implies that OthelloGPT knows which tiles are empty by $x^{0\_mid}$: after the first attention heads but before the MLP layer.
On a binary classification task of \textsc{Empty} vs. \textsc{Not-Empty} from 1,000 games in our test split, our probe achieves an accuracy of \textbf{76.8\%} and \textbf{98.9\%}, when projecting from $X^{0\_pre}$ and $x^{0\_mid}$ respectively.

\begin{figure}[t]
  \centering
    \includegraphics[clip, width=0.98\columnwidth]{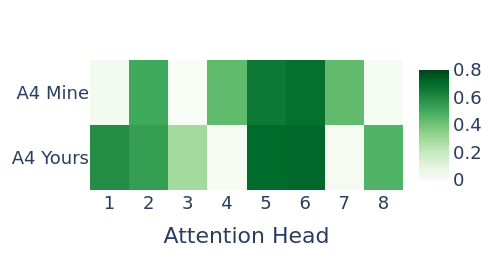}
  \caption{\label{fig:empty_logit_attribute}
Difference in probability of A4 being empty, between our clean and corrupt sequences, measured in each attention head.
}
\end{figure}

\begin{figure}[t]
  \centering
    \includegraphics[clip,  width=0.98\columnwidth]{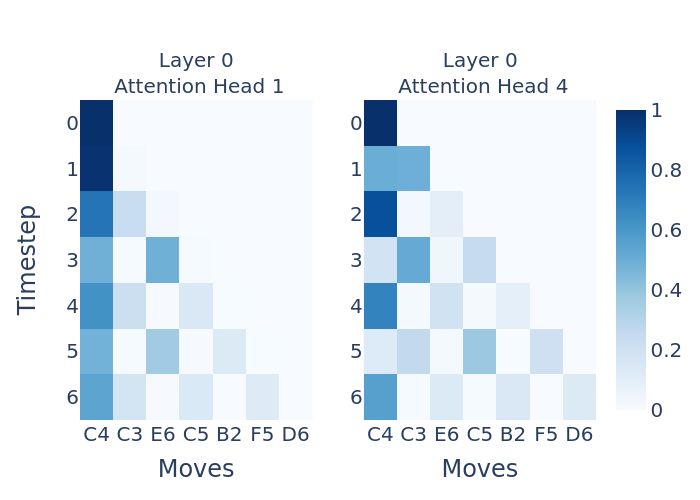}
\caption{\label{fig:attention_pattern}
Examples of attention heads attending to \textsc{Your} (left) or \textsc{My} (right) moves.
}
\end{figure}

\begin{table*}
\center
\begin{tabular}{ccccccccc}
\hline
    & $x^0$ & $x^1$ & $x^2$ & $x^3$ & $x^4$ & $x^5$ & $x^6$ & $x^7$  \\
\hline
Linear \{\textsc{Flipped}, \textsc{Not-Flipped}\} & 74.76 & 85.75 & 91.62 & 94.82 & 96.44 & 97.13 & 96.82 & 96.3 \\
\hline
\end{tabular}
\caption{$F1$ score for probing on \textsc{Flipped} tiles.
In addition to the board state, the model also linearly encodes concepts such as flipped tiles per timestep.
}
\label{tab:flipped_acc}
\end{table*}

\paragraph{Logit Attribute for \textsc{Empty}.}
The previous analysis is based on the \emph{weights} of the model.
Here we provide an alternative analysis by studying the \emph{activations} during inference.

First, we select a move $m$ (A4) that we wish to explain.
We then construct a ``clean'' and ``corrupt'' set of partial game sequences (N=4,569).
Our clean set always includes $m$, while our corrupt set replaces all timesteps with $m$ in the clean set with an alternative move.
We ensure that all games in our corrupt set remain legal sequences.
Finally, we study the \emph{difference in probability} that $m$ is empty, according to our probes, in our two sets.
Namely, we project the outputs from each attention head onto the \textsc{Empty} direction and apply a softmax:

\begin{align*}
    P_{\textsc{Empty}[m]}(\sigma) = Softmax(\sigma * p^\lambda_{\textsc{Empty}[m]})
\end{align*}

where $\sigma$ is the output from each attention head.

Figure~\ref{fig:empty_logit_attribute} shows the difference in probability that A4 is empty, between our clean and corrupt inputs, measured in each attention head of the first layer.
The figure decomposes two scenarios: when A4 was originally played by \textsc{Me} or \textsc{You}.
This is because some attention heads only attend to \textsc{My} moves (4, 7), while some only attend to \textsc{Yours} (1, 3, 8), which we show below.

\begin{figure*}[t]
  \centering
    \includegraphics[clip, width=0.98\textwidth]{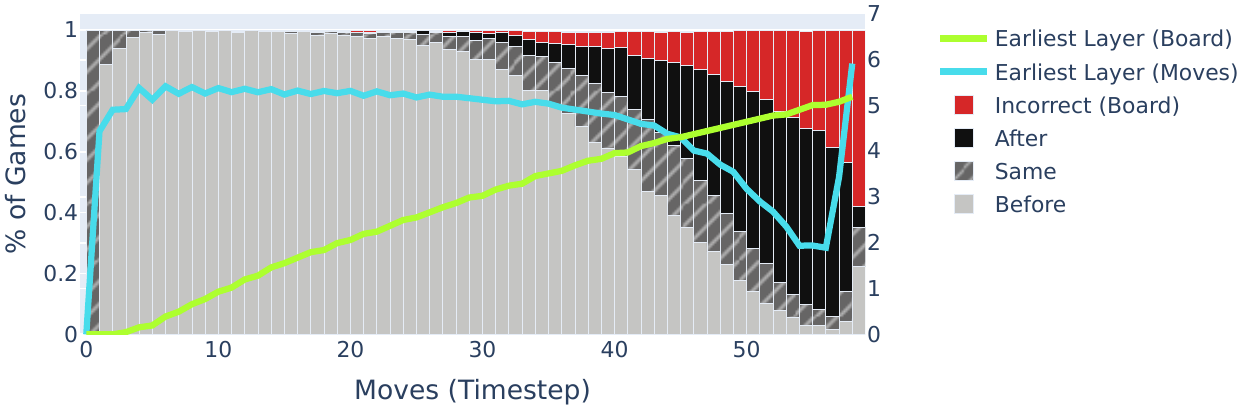}
  \caption{\label{fig:before_vs_after}
Proportion of times the board state is computed before/after move predictions are made (First y-axis).
\textbf{Light Grey:} Boards are computed in an earlier layer than moves.
\textbf{Dark Grey, Black:} Boards are computed in the same or later layer than moves.
\textbf{Red:} Model never computes the correct board state.
\textbf{Aqua, Lime (Curves):} Average earliest layer in which the board or moves are correctly computed (Second y-axis).
Starting from the mid-game, we start observing the model compute moves before boards (black bar), and this occurs more frequently as the game progresses.
}
\end{figure*}

\subsection{Attending to \textsc{My} \& \textsc{Your} Timesteps}
\label{subsec:attention_patterns}

We find that some attention heads only attend to either \textsc{My} or \textsc{Your} moves.
Figure~\ref{fig:attention_pattern} shows two examples: at each timestep, each head \emph{alternates} between attending to even or odd timesteps.
Such behavior further indicates how the model computes its world model based on \textsc{Mine} and \textsc{Yours} as opposed to \textsc{Black} and \textsc{White}.

\subsection{Additional Linear Concepts: \textsc{Flipped}}
\label{subsec:additional_linear_concepts}

In addition to linearly representing the board state, we find that OthelloGPT also encodes which tiles are being flipped, or captured, at each timestep.
To test this, we modify our probing task to classify between \textsc{Flipped} vs. \textsc{Not-Flipped}, with the same training setup described above.
Given the class imbalance, for this experiment we report $F1$ scores.
Table~\ref{tab:flipped_acc} demonstrates high $F1$ scores by layer 3.

We also conduct a modified version of our intervention experiment, in which we always randomly select a flipped tile at the current timestep to intervene on.
Then, instead of adding either $p^\lambda_{\textsc{Mine}}$,  $p^\lambda_{\textsc{Yours}}$, or $p^\lambda_{\textsc{Empty}}$, we \emph{subtract} $p^\lambda_{\textsc{Flipped}}$.
This tests whether the \textsc{Flipped} feature is causally relevant for computing the next move, by exploring whether this is sufficient to cause the model to play valid moves in the new board state. 
We get an average error rate of \textbf{0.486}, compared to a null intervention baseline rate of \textbf{1.686}.

One can consider \textsc{Flipped} tiles as the difference between the previous and current board state.
One might naturally think that a recurrent computation could derive the current board state by iteratively applying such differences.
However, transformer models do \textbf{not} make recursive computations!\footnote{Doing so would require our transformer model to have the same number of layers as the maximum game sequence length of 60.}
We view \textsc{Flipped} to be both an unexpected encoding and a hint for the rest of the board circuit.

\subsection{Multiple Circuits Hypothesis}
\label{subsec:multiple_circuits_hypothesis}

Although we find a board state circuit and its causality on move predictions, we find that it does not explain the entire model.
If our understanding is correct, we expect the model to compute the board state before computing valid moves.
However, we find that in end games, this is not the case.

To check for the correct board state, we apply our linear probes on each layer, and check the earliest layer in which all 64 tiles are correctly predicted.\footnote{It might be the case that legal moves could be predicted without 100\% accuracy of the board state. We try variants (see Appendix), but observe similar trends.}
To check for correct move predictions, we project from each layer using the unembedding layer, and check the earliest layer in which the top-N move predictions are all correct, where N is the number of groundtruth legal moves.

Figure~\ref{fig:before_vs_after} plots the proportion of times the board state is computed before (or after) valid moves (first y-axis).
We also overlay the average earliest layer in which board or moves are correctly computed (second y-axis, aqua and lime curves).
To our surprise, we find that in end games, the model often computes legal moves \emph{before} the board state (black bars).
We henceforth refer to this behavior as \textsc{MoveFirst}, and share some thoughts.

\paragraph{End Game Circuits.}
First, \textsc{MoveFirst} starts to occur around move 30, which is the mid-point of the game.
Second, \textsc{MoveFirst} occurs more frequently as we near the end of the game (increasing black bars).
Interestingly, in Othello, starting from the mid-point, there are progressively fewer empty tiles than there are filled tiles as the board fills up.
Also note that as the game progresses, it becomes more likely for every empty tile to be a legal move.

One possible explanation for this phenomenon is that in the end game, it may be possible to predict legal moves with simpler circuits that do not require the entire board state.
For instance, perhaps it combines \textsc{Empty} with other features such as \textsc{Is-Surrounded-By-Mine} or \textsc{Is-Border} and so on.

\paragraph{Multiple Circuits.}
Interestingly, the model still uses the board circuit at end games.
To demonstrate this, we run our intervention experiment on 1,000 \emph{end games},\footnote{We intervene on a timestep > 30} and still achieve a low error rate of \textbf{0.112}.\footnote{Non-intervention baseline: 1.988.}
We thus hypothesize that OthelloGPT (and more broadly, sequence models) consist of multiple circuits.
Another hypothesis is that residual networks make ``iterative inferences'' (Section~\ref{subsec:iterative_feature_refinement}), and for end games, OthelloGPT uses simpler circuits in the early layers and refines its predictions at late layers using board state.

\paragraph{End Game Board Accuracy.}
We observe that board state accuracy drops near end games.
This can be seen by the growing red bars, but also by measuring per-timestep accuracy of our probes (see Appendix).
It is unclear whether 1) the model does not bother to compute the perfect board state, as alternative circuits allow the model to still correctly predict legal moves, or 2) the model learns an alternative circuit because it struggles to compute the correct board state at end games.

\paragraph{Memorization.}
Note that in the first few timesteps, the board and legal moves are sometimes both computed in the same layer (dark grey bars).
This may be due to memorization: 1) these predictions both occur at the first layer, and 2) there are only so many openings in an Othello game.

\begin{figure*}[t]
  \centering
    \includegraphics[clip,  width=0.98\textwidth]{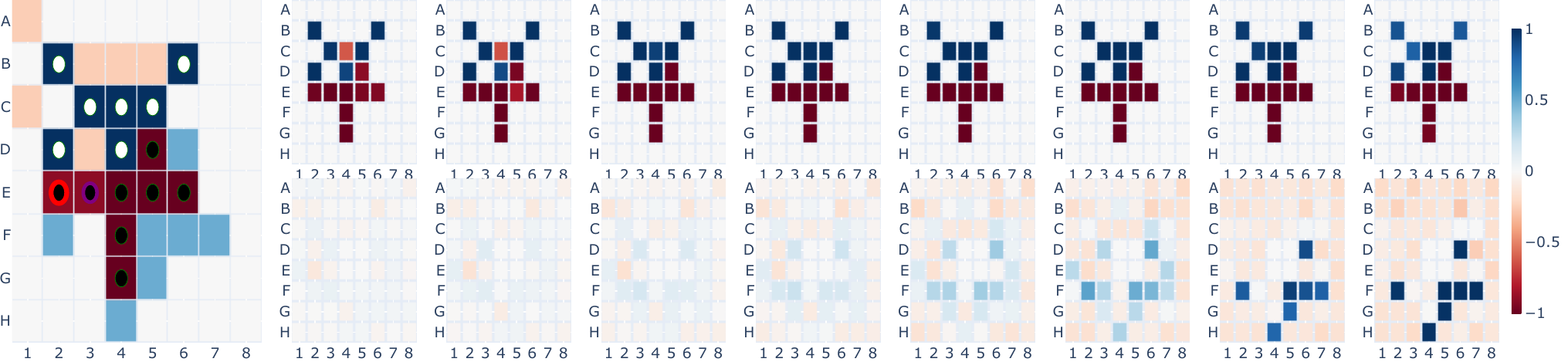}
  \caption{\label{fig:iterative_refinements}
Iterative refinements: the top row shows each layer projected using our linear probes.
The bottom row shows the model's predictions for legal moves at each layer, by applying the unembedding layer on each layer.
}
\end{figure*}

\subsection{Iterative Feature Refinements}
\label{subsec:iterative_feature_refinement}

Figure~\ref{fig:iterative_refinements} visualizes OthelloGPT's ``iterative inference'' \citep{jastrzebski2018residual, belrose2023eliciting, veit2016residual, logitlens}, or iterative refinement of features.
For each layer, we plot the projected board states using our probes, and projected next-move predictions using the unembedding layer.
Multiple evidence of iterative refinements are provided in the Appendix.

\section{Discussions}
\label{sec:discussion}

\subsection{On Linear vs. Non-Linear Interpretations}
\label{subsec:discussion_probing}

One challenge with probing is knowing which features to look for.\footnote{For a longer discussion on probing, see Appendix.}
For instance, classifying \{\textsc{Black}, \textsc{White}\} versus \{\textsc{Mine}, \textsc{Yours}\} leads to different takeaways, which illustrates the danger of \emph{projecting our preconceptions}.
What might seem ``sensible'' to a human interpreter (\textsc{Black, White}) may not be for a model.\footnote{In hindsight, given the symmetric game-play of Othello, encoding \textsc{Mine, Yours} is perfectly ``sensible'' for the model.}

More broadly, what is ``sensible'', or alternatively, how we choose to interpret linear or non-linear encodings, can be relative to how we see the world.
Suppose we had a perfect world model of our physical world.
Further suppose that if and when it computes a gravitational force between two objects (Newton's law), we discover a neuron whose square root was the distance between two objects.
Is this a non-linear representation of distance?
Or, given the form of Netwon's law, is the square of the distance a more natural way for the model to represent the feature, and thus considered a linear representation?
As this example shows, what constitutes a natural feature may be in the eye of the beholder.

\subsection{On the Emergence of Linear Representations}
\label{subsec:discussion_linear_emergence}

Linear representations in sequence models have been observed before: iGPT \citep{pmlr-v119-chen20s}, which was autoregressively trained to predict next pixels of images, lead to robust linear image representations.
The question remains, why do linear feature representations emerge?
What linear representations are currently encoded in large language models?
One reason might be simply that matrix multiplication can easily extract a different subset of linear features for each neuron.
However, we leave a complete explanation to future work.

\section{Related Work}
\label{sec:related_work}

We discuss three broad related areas: understanding internal representations, interventions, and mechanistic interpretability.

\subsection{Understanding Internal Representations}
\label{subsec:related_work:internal_reps}

Multiple researchers have studied world representations in sequence models.
\citet{li-etal-2021-implicit} train sequence models on a synthetic task, and uncover world models in their activations.
\citet{patel2022mapping} demonstrate that language models can learn to ground concepts (e.g., direction, colour) to real world representations.
\citet{burns2022dl} find linear vectors that encode ``truthfulness''.

Many studies also build or study linear representations for language.
Word embeddings \citep{NIPS2013_9aa42b31, mikolov2013efficient} build vectoral word representations.
Linear probes have also been used to extract linguistic characteristics in sentence embeddings \citep{conneau-etal-2018-cram, tenney-etal-2019-bert}.

Linear representations are found outside of language models as well.
\citet{merullo2022linearly} demonstrate that image representations from vision models can be linearly projected into the input space of language models.
\citet{alphazeroChess} and \citet{alphazeroHex} find interpretable representations of chess or Hex concepts in AlphaZero.

\subsection{Intervening On Language Models}
\label{subsec:related_work:intervene}

A growing body of work has intervened on language models, by which we mean controlling their behavior by altering their activations.

We consider two broad categories.
Parametric approaches often use optimizations (i.e. gradient descent) to locate and alter activations \citep{li2023emergent, meng2022locating, meng2022mass, hernandez2023measuring, hase2023does}.
Meanwhile, inference-time interventions typically apply linear arithmetics, for instance by using ``truthful'' vectors \citep{li2023inference}, ``task vectors'' \citep{ilharco2022editing}, or other ``steering vectors'' \citep{subramani-etal-2022-extracting, Turner_MacDiarmid_Udell_lisathiergart_Mini}.

\subsection{Mechanistic Interpretability}
\label{subsec:related_work:mech_int}

Mechanistic interpretability (MI) studies neural networks by reverse-engineering their behavior \citep{olah2020zoom, elhage2021mathematical}.
The goal of MI is to understand the underlying computations and representations of a model, with a broader goal of validating that their behavior aligns with what researchers have intended.
Such framework has allowed researchers to understand grokking \citep{nanda2023progress}, superposition \citep{elhage2022superposition, scherlis2022polysemanticity, arora2018linear}, or to study individual neurons \citep{mu2020compositional, antverg2021pitfalls, gurnee2023finding}.

\section{Conclusion}
\label{sec:conclusion}

In this work we demonstrated that the emergent world model in Othello-playing sequence models is full of linear representations.
Previously unbeknownst, we demonstrated that the board state in OthelloGPT is linearly represented by encoding the colour of each tile \emph{relative} to the player at each timestep (\textsc{Mine}, \textsc{Yours}, \textsc{Empty}) as opposed to absolute colour (\textsc{Black}, \textsc{White}, \textsc{Empty}).
We showed that we can accurately control the model's behaviour with simple vector arithmetic on the internal world model.
Lastly, we mechanistically interpreted multiple facets of the sequence model, analysing how empty tiles are detected, and linear representations of which pieces are flipped.
We find hints that multiple circuits might exist for predicting legal moves in the end game,
as well as further evidence that residual networks iteratively refine their features across layers.

\section{Acknowledgements}

We thank the original authors of \citet{li2023emergent} for opensourcing their work, making it possible to conduct our research.

We thank Chris Olah for invaluable discussion and encouragement, and drawing our attention to the implication of these results for the linear representation hypothesis. 

\section{Author Contributions}

Neel Nanda discovered the linear representation in terms of relative board state, and showed that simple vector arithmetic sufficed for causal interventions. He led an initial version of the experiments and write-ups, and advised throughout.

Andrew Lee led this write-up and performed all experiments in this paper. He discovered the flipped linear representation, the empty results, and the multiple circuit hypothesis results.

Martin Wattenberg helped with editing and distilling the paper, and contributed the analogy about a linear vs quadratic representation of distance.

\bibliography{custom}

\begin{thebibliography}{49}
\expandafter\ifx\csname natexlab\endcsname\relax\def\natexlab#1{#1}\fi

\bibitem[{Antverg and Belinkov(2021)}]{antverg2021pitfalls}
Omer Antverg and Yonatan Belinkov. 2021.
\newblock On the pitfalls of analyzing individual neurons in language models.
\newblock \emph{arXiv preprint arXiv:2110.07483}.

\bibitem[{Arora et~al.(2018)Arora, Li, Liang, Ma, and
  Risteski}]{arora2018linear}
Sanjeev Arora, Yuanzhi Li, Yingyu Liang, Tengyu Ma, and Andrej Risteski. 2018.
\newblock Linear algebraic structure of word senses, with applications to
  polysemy.
\newblock \emph{Transactions of the Association for Computational Linguistics},
  6:483--495.

\bibitem[{Bau et~al.(2020)Bau, Zhu, Strobelt, Lapedriza, Zhou, and
  Torralba}]{bau2020units}
David Bau, Jun-Yan Zhu, Hendrik Strobelt, Agata Lapedriza, Bolei Zhou, and
  Antonio Torralba. 2020.
\newblock \href {https://doi.org/10.1073/pnas.1907375117} {Understanding the
  role of individual units in a deep neural network}.
\newblock \emph{Proceedings of the National Academy of Sciences}.

\bibitem[{Belinkov(2022{\natexlab{a}})}]{10.1162/coli_a_00422}
Yonatan Belinkov. 2022{\natexlab{a}}.
\newblock \href {https://doi.org/10.1162/coli_a_00422} {Probing classifiers:
  Promises, shortcomings, and advances}.
\newblock \emph{Computational Linguistics}, 48(1):207--219.

\bibitem[{Belinkov(2022{\natexlab{b}})}]{belinkov-2022-probing}
Yonatan Belinkov. 2022{\natexlab{b}}.
\newblock \href {https://doi.org/10.1162/coli_a_00422} {Probing classifiers:
  Promises, shortcomings, and advances}.
\newblock \emph{Computational Linguistics}, 48(1):207--219.

\bibitem[{Belrose et~al.(2023)Belrose, Furman, Smith, Halawi, Ostrovsky,
  McKinney, Biderman, and Steinhardt}]{belrose2023eliciting}
Nora Belrose, Zach Furman, Logan Smith, Danny Halawi, Igor Ostrovsky, Lev
  McKinney, Stella Biderman, and Jacob Steinhardt. 2023.
\newblock Eliciting latent predictions from transformers with the tuned lens.
\newblock \emph{arXiv preprint arXiv:2303.08112}.

\bibitem[{Burns et~al.(2022)Burns, Ye, Klein, and Steinhardt}]{burns2022dl}
Collin Burns, Haotian Ye, Dan Klein, and Jacob Steinhardt. 2022.
\newblock Discovering latent knowledge in language models without supervision.
\newblock \emph{ArXiV}.

\bibitem[{Chen et~al.(2020)Chen, Radford, Child, Wu, Jun, Luan, and
  Sutskever}]{pmlr-v119-chen20s}
Mark Chen, Alec Radford, Rewon Child, Jeffrey Wu, Heewoo Jun, David Luan, and
  Ilya Sutskever. 2020.
\newblock \href {https://proceedings.mlr.press/v119/chen20s.html} {Generative
  pretraining from pixels}.
\newblock In \emph{Proceedings of the 37th International Conference on Machine
  Learning}, volume 119 of \emph{Proceedings of Machine Learning Research},
  pages 1691--1703. PMLR.

\bibitem[{Conneau et~al.(2018)Conneau, Kruszewski, Lample, Barrault, and
  Baroni}]{conneau-etal-2018-cram}
Alexis Conneau, German Kruszewski, Guillaume Lample, Lo{\"\i}c Barrault, and
  Marco Baroni. 2018.
\newblock \href {https://doi.org/10.18653/v1/P18-1198} {What you can cram into
  a single {\$}{\&}!{\#}* vector: Probing sentence embeddings for linguistic
  properties}.
\newblock In \emph{Proceedings of the 56th Annual Meeting of the Association
  for Computational Linguistics (Volume 1: Long Papers)}, pages 2126--2136,
  Melbourne, Australia. Association for Computational Linguistics.

\bibitem[{Elhage et~al.(2022{\natexlab{a}})Elhage, Hume, Olsson, Nanda,
  Henighan, Johnston, ElShowk, Joseph, DasSarma, Mann, Hernandez, Askell,
  Ndousse, Jones, Drain, Chen, Bai, Ganguli, Lovitt, Hatfield-Dodds, Kernion,
  Conerly, Kravec, Fort, Kadavath, Jacobson, Tran-Johnson, Kaplan, Clark,
  Brown, McCandlish, Amodei, and Olah}]{elhage2022solu}
Nelson Elhage, Tristan Hume, Catherine Olsson, Neel Nanda, Tom Henighan, Scott
  Johnston, Sheer ElShowk, Nicholas Joseph, Nova DasSarma, Ben Mann, Danny
  Hernandez, Amanda Askell, Kamal Ndousse, Andy Jones, Dawn Drain, Anna Chen,
  Yuntao Bai, Deep Ganguli, Liane Lovitt, Zac Hatfield-Dodds, Jackson Kernion,
  Tom Conerly, Shauna Kravec, Stanislav Fort, Saurav Kadavath, Josh Jacobson,
  Eli Tran-Johnson, Jared Kaplan, Jack Clark, Tom Brown, Sam McCandlish, Dario
  Amodei, and Christopher Olah. 2022{\natexlab{a}}.
\newblock Softmax linear units.
\newblock \emph{Transformer Circuits Thread}.
\newblock Https://transformer-circuits.pub/2022/solu/index.html.

\bibitem[{Elhage et~al.(2022{\natexlab{b}})Elhage, Hume, Olsson, Schiefer,
  Henighan, Kravec, Hatfield-Dodds, Lasenby, Drain, Chen, Grosse, McCandlish,
  Kaplan, Amodei, Wattenberg, and Olah}]{elhage2022superposition}
Nelson Elhage, Tristan Hume, Catherine Olsson, Nicholas Schiefer, Tom Henighan,
  Shauna Kravec, Zac Hatfield-Dodds, Robert Lasenby, Dawn Drain, Carol Chen,
  Roger Grosse, Sam McCandlish, Jared Kaplan, Dario Amodei, Martin Wattenberg,
  and Christopher Olah. 2022{\natexlab{b}}.
\newblock \href {https://transformer-circuits.pub/2022/toy_model/index.html}
  {Toy models of superposition}.
\newblock \emph{Transformer Circuits Thread}.

\bibitem[{Elhage et~al.(2021)Elhage, Nanda, Olsson, Henighan, Joseph, Mann,
  Askell, Bai, Chen, Conerly, DasSarma, Drain, Ganguli, Hatfield-Dodds,
  Hernandez, Jones, Kernion, Lovitt, Ndousse, Amodei, Brown, Clark, Kaplan,
  McCandlish, and Olah}]{elhage2021mathematical}
Nelson Elhage, Neel Nanda, Catherine Olsson, Tom Henighan, Nicholas Joseph, Ben
  Mann, Amanda Askell, Yuntao Bai, Anna Chen, Tom Conerly, Nova DasSarma, Dawn
  Drain, Deep Ganguli, Zac Hatfield-Dodds, Danny Hernandez, Andy Jones, Jackson
  Kernion, Liane Lovitt, Kamal Ndousse, Dario Amodei, Tom Brown, Jack Clark,
  Jared Kaplan, Sam McCandlish, and Chris Olah. 2021.
\newblock A mathematical framework for transformer circuits.
\newblock \emph{Transformer Circuits Thread}.
\newblock Https://transformer-circuits.pub/2021/framework/index.html.

\bibitem[{Geva et~al.(2021)Geva, Schuster, Berant, and
  Levy}]{geva-etal-2021-transformer}
Mor Geva, Roei Schuster, Jonathan Berant, and Omer Levy. 2021.
\newblock \href {https://doi.org/10.18653/v1/2021.emnlp-main.446} {Transformer
  feed-forward layers are key-value memories}.
\newblock In \emph{Proceedings of the 2021 Conference on Empirical Methods in
  Natural Language Processing}, pages 5484--5495, Online and Punta Cana,
  Dominican Republic. Association for Computational Linguistics.

\bibitem[{Giulianelli et~al.(2018)Giulianelli, Harding, Mohnert, Hupkes, and
  Zuidema}]{giulianelli-etal-2018-hood}
Mario Giulianelli, Jack Harding, Florian Mohnert, Dieuwke Hupkes, and Willem
  Zuidema. 2018.
\newblock \href {https://doi.org/10.18653/v1/W18-5426} {Under the hood: Using
  diagnostic classifiers to investigate and improve how language models track
  agreement information}.
\newblock In \emph{Proceedings of the 2018 {EMNLP} Workshop {B}lackbox{NLP}:
  Analyzing and Interpreting Neural Networks for {NLP}}, pages 240--248,
  Brussels, Belgium. Association for Computational Linguistics.

\bibitem[{Goh et~al.(2021)Goh, †, †, Carter, Petrov, Schubert, Radford, and
  Olah}]{goh2021multimodal}
Gabriel Goh, Nick~Cammarata †, Chelsea~Voss †, Shan Carter, Michael Petrov,
  Ludwig Schubert, Alec Radford, and Chris Olah. 2021.
\newblock \href {https://doi.org/10.23915/distill.00030} {Multimodal neurons in
  artificial neural networks}.
\newblock \emph{Distill}.
\newblock Https://distill.pub/2021/multimodal-neurons.

\bibitem[{Gurnee et~al.(2023)Gurnee, Nanda, Pauly, Harvey, Troitskii, and
  Bertsimas}]{gurnee2023finding}
Wes Gurnee, Neel Nanda, Matthew Pauly, Katherine Harvey, Dmitrii Troitskii, and
  Dimitris Bertsimas. 2023.
\newblock Finding neurons in a haystack: Case studies with sparse probing.
\newblock \emph{arXiv preprint arXiv:2305.01610}.

\bibitem[{Hase et~al.(2023)Hase, Bansal, Kim, and Ghandeharioun}]{hase2023does}
Peter Hase, Mohit Bansal, Been Kim, and Asma Ghandeharioun. 2023.
\newblock Does localization inform editing? surprising differences in
  causality-based localization vs. knowledge editing in language models.
\newblock \emph{arXiv preprint arXiv:2301.04213}.

\bibitem[{Hernandez et~al.(2023)Hernandez, Li, and
  Andreas}]{hernandez2023measuring}
Evan Hernandez, Belinda~Z Li, and Jacob Andreas. 2023.
\newblock Measuring and manipulating knowledge representations in language
  models.
\newblock \emph{arXiv preprint arXiv:2304.00740}.

\bibitem[{Ilharco et~al.(2022)Ilharco, Ribeiro, Wortsman, Gururangan, Schmidt,
  Hajishirzi, and Farhadi}]{ilharco2022editing}
Gabriel Ilharco, Marco~Tulio Ribeiro, Mitchell Wortsman, Suchin Gururangan,
  Ludwig Schmidt, Hannaneh Hajishirzi, and Ali Farhadi. 2022.
\newblock Editing models with task arithmetic.
\newblock \emph{arXiv preprint arXiv:2212.04089}.

\bibitem[{Jastrzebski et~al.(2018)Jastrzebski, Arpit, Ballas, Verma, Che, and
  Bengio}]{jastrzebski2018residual}
Stanisław Jastrzebski, Devansh Arpit, Nicolas Ballas, Vikas Verma, Tong Che,
  and Yoshua Bengio. 2018.
\newblock \href {https://openreview.net/forum?id=SJa9iHgAZ} {Residual
  connections encourage iterative inference}.
\newblock In \emph{International Conference on Learning Representations}.

\bibitem[{Li et~al.(2021)Li, Nye, and Andreas}]{li-etal-2021-implicit}
Belinda~Z. Li, Maxwell Nye, and Jacob Andreas. 2021.
\newblock \href {https://doi.org/10.18653/v1/2021.acl-long.143} {Implicit
  representations of meaning in neural language models}.
\newblock In \emph{Proceedings of the 59th Annual Meeting of the Association
  for Computational Linguistics and the 11th International Joint Conference on
  Natural Language Processing (Volume 1: Long Papers)}, pages 1813--1827,
  Online. Association for Computational Linguistics.

\bibitem[{Li et~al.(2023{\natexlab{a}})Li, Hopkins, Bau, Vi{\'e}gas, Pfister,
  and Wattenberg}]{li2023emergent}
Kenneth Li, Aspen~K Hopkins, David Bau, Fernanda Vi{\'e}gas, Hanspeter Pfister,
  and Martin Wattenberg. 2023{\natexlab{a}}.
\newblock \href {https://openreview.net/forum?id=DeG07_TcZvT} {Emergent world
  representations: Exploring a sequence model trained on a synthetic task}.
\newblock In \emph{The Eleventh International Conference on Learning
  Representations}.

\bibitem[{Li et~al.(2023{\natexlab{b}})Li, Patel, Vi{\'e}gas, Pfister, and
  Wattenberg}]{li2023inference}
Kenneth Li, Oam Patel, Fernanda Vi{\'e}gas, Hanspeter Pfister, and Martin
  Wattenberg. 2023{\natexlab{b}}.
\newblock Inference-time intervention: Eliciting truthful answers from a
  language model.
\newblock \emph{arXiv preprint arXiv:2306.03341}.

\bibitem[{Lovering et~al.(2022)Lovering, Forde, Konidaris, Pavlick, and
  Littman}]{alphazeroHex}
Charles Lovering, Jessica Forde, George Konidaris, Ellie Pavlick, and Michael
  Littman. 2022.
\newblock \href
  {https://proceedings.neurips.cc/paper_files/paper/2022/file/a705747417d32ebf1916169e1a442274-Paper-Conference.pdf}
  {Evaluation beyond task performance: Analyzing concepts in alphazero in hex}.
\newblock In \emph{Advances in Neural Information Processing Systems},
  volume~35, pages 25992--26006. Curran Associates, Inc.

\bibitem[{McGrath et~al.(2022)McGrath, Kapishnikov, Tomašev, Pearce,
  Wattenberg, Hassabis, Kim, Paquet, and Kramnik}]{alphazeroChess}
Thomas McGrath, Andrei Kapishnikov, Nenad Tomašev, Adam Pearce, Martin
  Wattenberg, Demis Hassabis, Been Kim, Ulrich Paquet, and Vladimir Kramnik.
  2022.
\newblock \href {https://doi.org/10.1073/pnas.2206625119} {Acquisition of chess
  knowledge in alphazero}.
\newblock \emph{Proceedings of the National Academy of Sciences},
  119(47):e2206625119.

\bibitem[{McGrath et~al.(2023)McGrath, Rahtz, Kramar, Mikulik, and
  Legg}]{mcgrath2023hydra}
Thomas McGrath, Matthew Rahtz, Janos Kramar, Vladimir Mikulik, and Shane Legg.
  2023.
\newblock The hydra effect: Emergent self-repair in language model
  computations.
\newblock \emph{arXiv preprint arXiv:2307.15771}.

\bibitem[{Meng et~al.(2022{\natexlab{a}})Meng, Bau, Andonian, and
  Belinkov}]{meng2022locating}
Kevin Meng, David Bau, Alex Andonian, and Yonatan Belinkov. 2022{\natexlab{a}}.
\newblock Locating and editing factual associations in {GPT}.
\newblock \emph{Advances in Neural Information Processing Systems}, 36.

\bibitem[{Meng et~al.(2022{\natexlab{b}})Meng, Sharma, Andonian, Belinkov, and
  Bau}]{meng2022mass}
Kevin Meng, Arnab~Sen Sharma, Alex Andonian, Yonatan Belinkov, and David Bau.
  2022{\natexlab{b}}.
\newblock Mass-editing memory in a transformer.
\newblock \emph{arXiv preprint arXiv:2210.07229}.

\bibitem[{Merullo et~al.(2022)Merullo, Castricato, Eickhoff, and
  Pavlick}]{merullo2022linearly}
Jack Merullo, Louis Castricato, Carsten Eickhoff, and Ellie Pavlick. 2022.
\newblock Linearly mapping from image to text space.
\newblock \emph{arXiv preprint arXiv:2209.15162}.

\bibitem[{Mikolov et~al.(2013{\natexlab{a}})Mikolov, Chen, Corrado, and
  Dean}]{mikolov2013efficient}
Tomas Mikolov, Kai Chen, Greg Corrado, and Jeffrey Dean. 2013{\natexlab{a}}.
\newblock Efficient estimation of word representations in vector space.
\newblock \emph{arXiv preprint arXiv:1301.3781}.

\bibitem[{Mikolov et~al.(2013{\natexlab{b}})Mikolov, Sutskever, Chen, Corrado,
  and Dean}]{NIPS2013_9aa42b31}
Tomas Mikolov, Ilya Sutskever, Kai Chen, Greg~S Corrado, and Jeff Dean.
  2013{\natexlab{b}}.
\newblock \href
  {https://proceedings.neurips.cc/paper_files/paper/2013/file/9aa42b31882ec039965f3c4923ce901b-Paper.pdf}
  {Distributed representations of words and phrases and their
  compositionality}.
\newblock In \emph{Advances in Neural Information Processing Systems},
  volume~26. Curran Associates, Inc.

\bibitem[{Mikolov et~al.(2013{\natexlab{c}})Mikolov, Yih, and
  Zweig}]{mikolov2013linguistic}
Tom{\'a}{\v{s}} Mikolov, Wen-tau Yih, and Geoffrey Zweig. 2013{\natexlab{c}}.
\newblock Linguistic regularities in continuous space word representations.
\newblock In \emph{Proceedings of the 2013 conference of the north american
  chapter of the association for computational linguistics: Human language
  technologies}, pages 746--751.

\bibitem[{Mu and Andreas(2020)}]{mu2020compositional}
Jesse Mu and Jacob Andreas. 2020.
\newblock Compositional explanations of neurons.
\newblock \emph{Advances in Neural Information Processing Systems},
  33:17153--17163.

\bibitem[{Nanda et~al.(2023)Nanda, Chan, Liberum, Smith, and
  Steinhardt}]{nanda2023progress}
Neel Nanda, Lawrence Chan, Tom Liberum, Jess Smith, and Jacob Steinhardt. 2023.
\newblock Progress measures for grokking via mechanistic interpretability.
\newblock \emph{arXiv preprint arXiv:2301.05217}.

\bibitem[{nostalgebraist(2020)}]{logitlens}
nostalgebraist. 2020.
\newblock \href
  {https://www.lesswrong.com/posts/AcKRB8wDpdaN6v6ru/interpreting-gpt-the-logit-lens}
  {interpreting gpt: the logit lens}.

\bibitem[{Olah et~al.(2020)Olah, Cammarata, Schubert, Goh, Petrov, and
  Carter}]{olah2020zoom}
Chris Olah, Nick Cammarata, Ludwig Schubert, Gabriel Goh, Michael Petrov, and
  Shan Carter. 2020.
\newblock \href {https://doi.org/10.23915/distill.00024.001} {Zoom in: An
  introduction to circuits}.
\newblock \emph{Distill}.
\newblock Https://distill.pub/2020/circuits/zoom-in.

\bibitem[{Patel and Pavlick(2022)}]{patel2022mapping}
Roma Patel and Ellie Pavlick. 2022.
\newblock Mapping language models to grounded conceptual spaces.
\newblock In \emph{International Conference on Learning Representations}.

\bibitem[{Pimentel et~al.(2020{\natexlab{a}})Pimentel, Saphra, Williams, and
  Cotterell}]{pimentel-etal-2020-pareto}
Tiago Pimentel, Naomi Saphra, Adina Williams, and Ryan Cotterell.
  2020{\natexlab{a}}.
\newblock \href {https://doi.org/10.18653/v1/2020.emnlp-main.254} {{P}areto
  probing: {T}rading off accuracy for complexity}.
\newblock In \emph{Proceedings of the 2020 Conference on Empirical Methods in
  Natural Language Processing (EMNLP)}, pages 3138--3153, Online. Association
  for Computational Linguistics.

\bibitem[{Pimentel et~al.(2020{\natexlab{b}})Pimentel, Valvoda, Maudslay,
  Zmigrod, Williams, and Cotterell}]{pimentel-etal-2020-information}
Tiago Pimentel, Josef Valvoda, Rowan~Hall Maudslay, Ran Zmigrod, Adina
  Williams, and Ryan Cotterell. 2020{\natexlab{b}}.
\newblock \href {https://doi.org/10.18653/v1/2020.acl-main.420}
  {Information-theoretic probing for linguistic structure}.
\newblock In \emph{Proceedings of the 58th Annual Meeting of the Association
  for Computational Linguistics}, pages 4609--4622, Online. Association for
  Computational Linguistics.

\bibitem[{Radford et~al.(2019)Radford, Wu, Child, Luan, Amodei, Sutskever
  et~al.}]{radford2019language}
Alec Radford, Jeffrey Wu, Rewon Child, David Luan, Dario Amodei, Ilya
  Sutskever, et~al. 2019.
\newblock Language models are unsupervised multitask learners.

\bibitem[{Saphra and Lopez(2019)}]{saphra-lopez-2019-understanding}
Naomi Saphra and Adam Lopez. 2019.
\newblock \href {https://doi.org/10.18653/v1/N19-1329} {Understanding learning
  dynamics of language models with {SVCCA}}.
\newblock In \emph{Proceedings of the 2019 Conference of the North {A}merican
  Chapter of the Association for Computational Linguistics: Human Language
  Technologies, Volume 1 (Long and Short Papers)}, pages 3257--3267,
  Minneapolis, Minnesota. Association for Computational Linguistics.

\bibitem[{Scherlis et~al.(2022)Scherlis, Sachan, Jermyn, Benton, and
  Shlegeris}]{scherlis2022polysemanticity}
Adam Scherlis, Kshitij Sachan, Adam~S Jermyn, Joe Benton, and Buck Shlegeris.
  2022.
\newblock Polysemanticity and capacity in neural networks.
\newblock \emph{arXiv preprint arXiv:2210.01892}.

\bibitem[{Silver et~al.(2018)Silver, Hubert, Schrittwieser, Antonoglou, Lai,
  Guez, Lanctot, Sifre, Kumaran, Graepel, Lillicrap, Simonyan, and
  Hassabis}]{alphazero}
David Silver, Thomas Hubert, Julian Schrittwieser, Ioannis Antonoglou, Matthew
  Lai, Arthur Guez, Marc Lanctot, Laurent Sifre, Dharshan Kumaran, Thore
  Graepel, Timothy Lillicrap, Karen Simonyan, and Demis Hassabis. 2018.
\newblock \href {https://doi.org/10.1126/science.aar6404} {A general
  reinforcement learning algorithm that masters chess, shogi, and go through
  self-play}.
\newblock \emph{Science}, 362(6419):1140--1144.

\bibitem[{Subramani et~al.(2022)Subramani, Suresh, and
  Peters}]{subramani-etal-2022-extracting}
Nishant Subramani, Nivedita Suresh, and Matthew Peters. 2022.
\newblock \href {https://doi.org/10.18653/v1/2022.findings-acl.48} {Extracting
  latent steering vectors from pretrained language models}.
\newblock In \emph{Findings of the Association for Computational Linguistics:
  ACL 2022}, pages 566--581, Dublin, Ireland. Association for Computational
  Linguistics.

\bibitem[{Tenney et~al.(2019)Tenney, Das, and Pavlick}]{tenney-etal-2019-bert}
Ian Tenney, Dipanjan Das, and Ellie Pavlick. 2019.
\newblock \href {https://doi.org/10.18653/v1/P19-1452} {{BERT} rediscovers the
  classical {NLP} pipeline}.
\newblock In \emph{Proceedings of the 57th Annual Meeting of the Association
  for Computational Linguistics}, pages 4593--4601, Florence, Italy.
  Association for Computational Linguistics.

\bibitem[{Tucker et~al.(2021)Tucker, Qian, and
  Levy}]{tucker-etal-2021-modified}
Mycal Tucker, Peng Qian, and Roger Levy. 2021.
\newblock \href {https://doi.org/10.18653/v1/2021.findings-acl.76} {What if
  this modified that? syntactic interventions with counterfactual embeddings}.
\newblock In \emph{Findings of the Association for Computational Linguistics:
  ACL-IJCNLP 2021}, pages 862--875, Online. Association for Computational
  Linguistics.

\bibitem[{Turner et~al.(2023)Turner, MacDiarmid, Udell, lisathiergart, and
  Mini}]{Turner_MacDiarmid_Udell_lisathiergart_Mini}
Alex Turner, Monte MacDiarmid, David Udell, lisathiergart, and Ulisse Mini.
  2023.
\newblock \href
  {https://www.alignmentforum.org/posts/5spBue2z2tw4JuDCx/steering-gpt-2-xl-by-adding-an-activation-vector}
  {Steering gpt-2-xl by adding an activation vector - ai alignment forum}.

\bibitem[{Vaswani et~al.(2017)Vaswani, Shazeer, Parmar, Uszkoreit, Jones,
  Gomez, Kaiser, and Polosukhin}]{NIPS2017_3f5ee243}
Ashish Vaswani, Noam Shazeer, Niki Parmar, Jakob Uszkoreit, Llion Jones,
  Aidan~N Gomez, \L~ukasz Kaiser, and Illia Polosukhin. 2017.
\newblock \href
  {https://proceedings.neurips.cc/paper_files/paper/2017/file/3f5ee243547dee91fbd053c1c4a845aa-Paper.pdf}
  {Attention is all you need}.
\newblock In \emph{Advances in Neural Information Processing Systems},
  volume~30. Curran Associates, Inc.

\bibitem[{Veit et~al.(2016)Veit, Wilber, and Belongie}]{veit2016residual}
Andreas Veit, Michael~J Wilber, and Serge Belongie. 2016.
\newblock Residual networks behave like ensembles of relatively shallow
  networks.
\newblock \emph{Advances in neural information processing systems}, 29.

\end{thebibliography}
\bibliographystyle{acl_natbib}

\clearpage
\appendix

\clearpage
\pdfoutput=1








\appendix

\begin{table}
\centering
\small
\begin{tabular}{l|c}
    \toprule
    Hyperparameter  &  Value  \\
    \midrule
    Optimizer       & AdamW \\
    Learning Rate   & 1e-2 \\
    Weight Decay    & 1e-2 \\
    Betas           & 0.9, 0.99 \\
    Validation Step & 200 \\
    Validation Size & 512 \\
    Validation Patience & 10 \\
    \bottomrule
\end{tabular}

\caption{\label{tab:appx_hyperparams}
Hyperparameters used for our linear probes.}
\end{table} 

\section{Hyperparameters for Linear Probes}
\label{sec:appx_hyperparams}

Table~\ref{tab:appx_hyperparams} provides hyperparameters used for our linear probes.

\begin{figure}[t]
  \centering
    \includegraphics[trim={0cm 0cm 2cm 0cm}, clip, width=0.98\columnwidth]{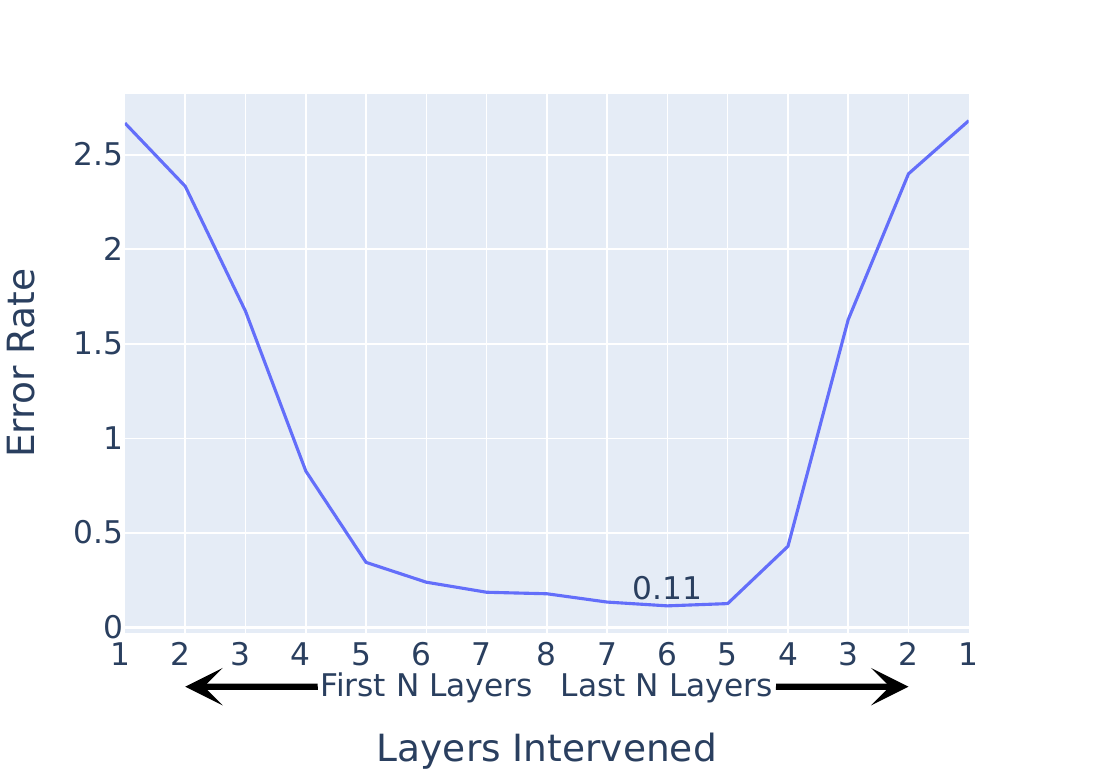}
  \caption{\label{fig:appx_intervene_layers}
Intervention results depending on layers intervened.
}
\end{figure}

\section{Intervening on Different Layers}
\label{sec:appx_intervene_diff_layers}

In practice there are a lot of ways to intervene using linear vectors.
Figure~\ref{fig:appx_intervene_layers} demonstrates different error rates depending on which layers are intervened.
From our experiments, we observe that either a sufficient number of layers need to be intervened for OthelloGPT to alter its predictions.
We offer a couple of hypotheses for this.
First, we hypothesize that this is because of the residual structure of transformer models, and while each layer may write additional information into the residual streams, there may still be information from earlier layers that the model uses.
A somewhat related hypothesis is that OthelloGPT might be demonstrating the Hydra effect \citep{mcgrath2023hydra}, in which language models demonstrate the ability to self-repair its computations after an intervention.

\begin{figure*}[t]
  \centering
    \includegraphics[clip, width=0.98\textwidth]{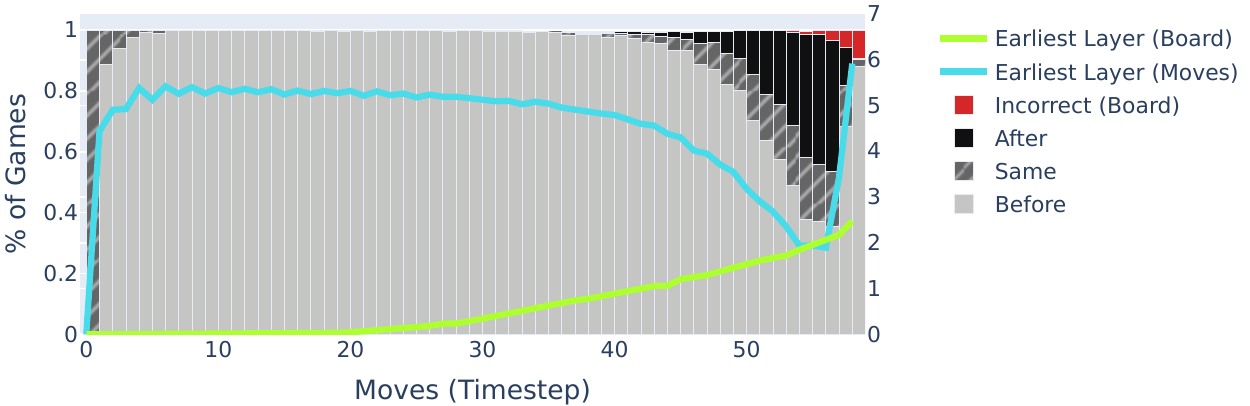}
  \caption{\label{fig:before_vs_after_0.9}
Percentage of times \textbf{90\%} of the board state is computed before/after move predictions are made.}
\end{figure*}

\begin{figure*}[t]
  \centering
    \includegraphics[clip, width=0.98\textwidth]{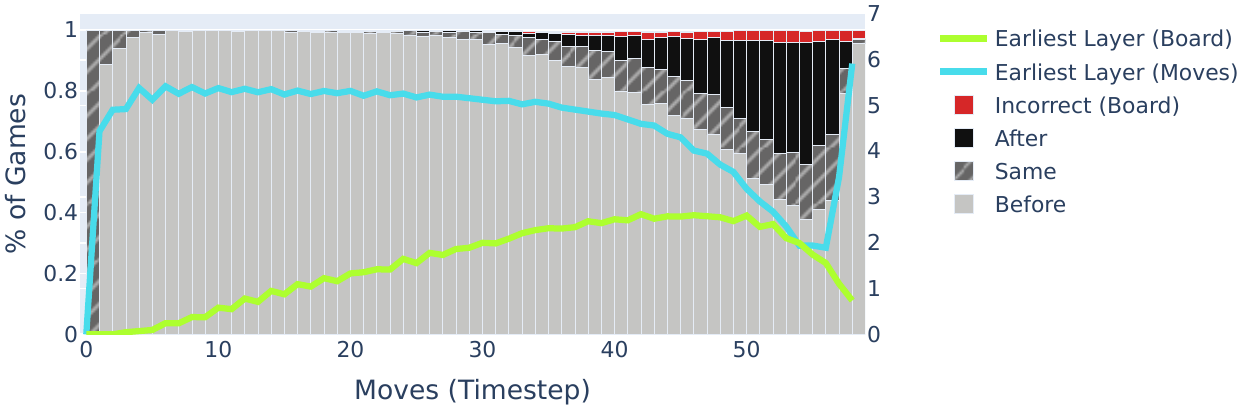}
  \caption{\label{fig:before_vs_after_skeleton}
Percentage of times the \textbf{``minimum set''} of necessary board state is computed before/after move predictions are made.}
\end{figure*}

\section{Multiple Circuits}
\label{sec:appx_multiple_cuircuits}

\begin{figure}[t]
  \centering
    \includegraphics[clip,  width=0.98\columnwidth]{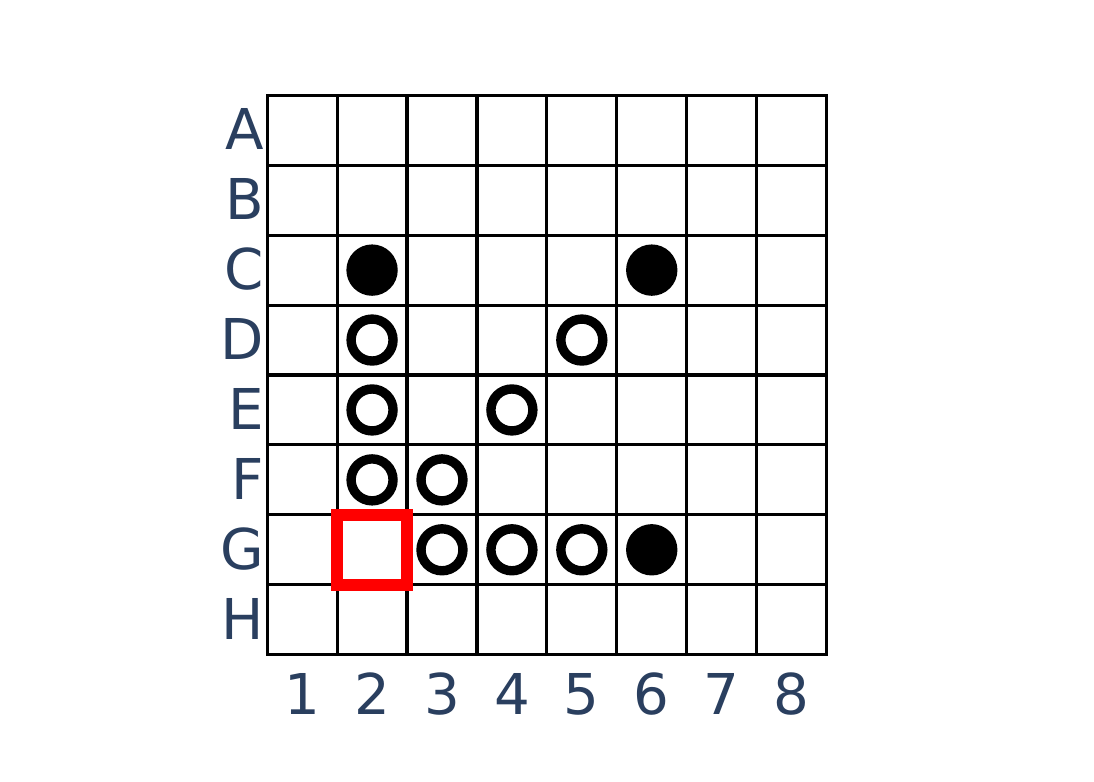}
  \caption{\label{fig:appx_minimum_tiles}
Example of ``minimum set'' of tiles that make move G2 legal. 
}
\end{figure}

In Section~\ref{subsec:multiple_circuits_hypothesis}, we find hints that OthelloGPT sometimes computes moves before boards at end games.
Namely, we check the earliest layers in which the board is correctly predicted with 100\% accuracy.
Could it be that at end games, legal moves can be predicted without needing the entire board? 
To this point, we experiment with variations of this experiment.
In Figure~\ref{fig:before_vs_after_0.9}, we check the earliest layer in which at least 90\% of the board is first correctly computed.
In Figure~\ref{fig:before_vs_after_skeleton}, we check the earliest layer in which the ``minimum set'' of tiles are correctly computed, where the minimum set is set of tiles that make each legal move playable (see Figure~\ref{fig:appx_minimum_tiles} for example).
Despite a looser criteria for board state, we still see OthelloGPT computing moves before boards at end games.

\begin{figure}[t]
  \centering
    \includegraphics[clip,  width=0.98\columnwidth]{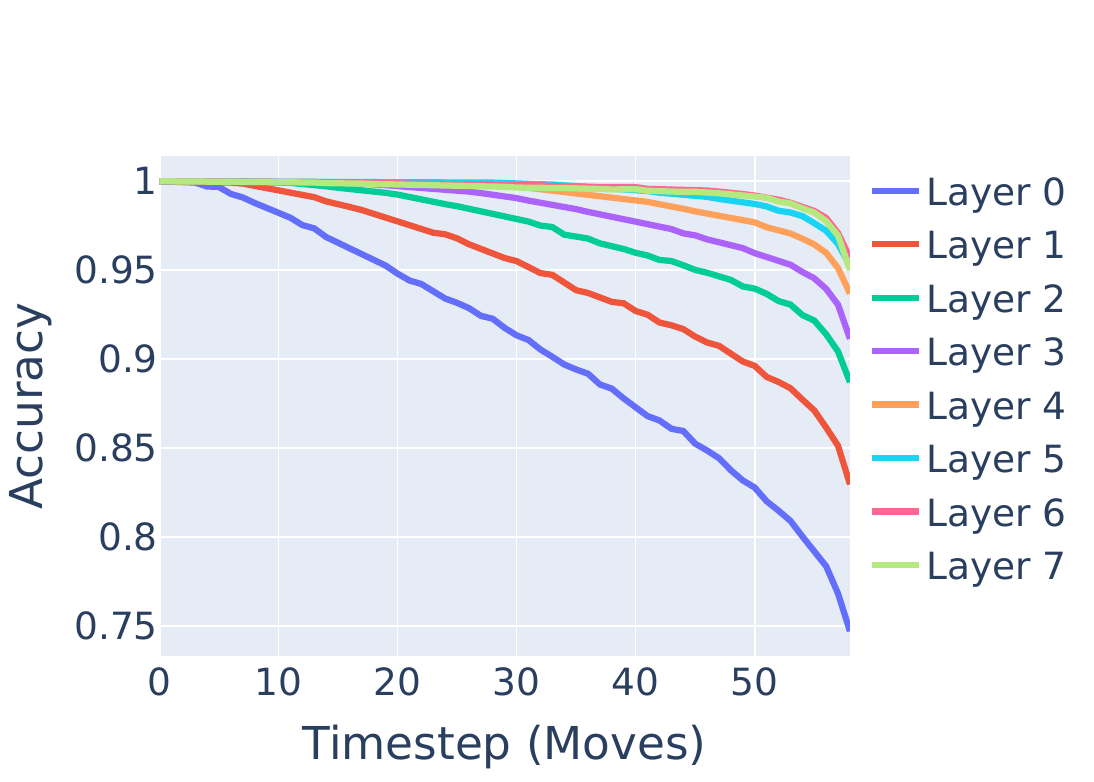}
  \caption{\label{fig:appx_per_timestep_accuracy}
Accuracy per timestep for our linear probes.
}
\end{figure}

Interestingly, our probes lose accuracy starts to drop in the end game as well (Figure~\ref{fig:appx_per_timestep_accuracy}).
It is unclear whether 1) the model does not bother to compute the perfect board state, as alternative circuits might exist at end games, or 2) the model learns an alternative circuit because it struggles to compute the correct board state at end games.

\section{Evidence of Iterative Feature Refinements}
\label{sec:appx_additional_evidence_iterative_features}

\begin{figure}[t]
  \centering
    \includegraphics[clip,  width=0.98\columnwidth]{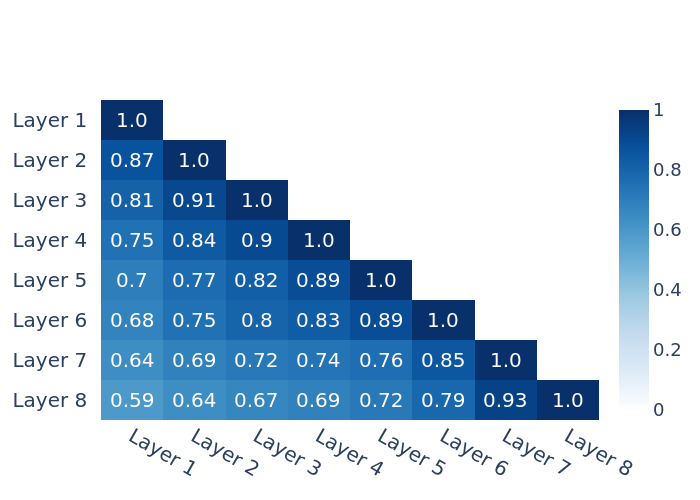}
  \caption{\label{fig:cos_sims}
Cosine similarity scores between linear probes across layers.
}
\end{figure}

As mentioned in Section~\ref{subsec:iterative_feature_refinement}, OthelloGPT demonstrates multiple evidence of iterative feature refinements:
1) Board state accuracy (as well as \textsc{Flipped}) improves from layer to layer (Table~\ref{tab:linear_acc}, ~\ref{tab:flipped_acc}).
2) Next-move predictions also improve from layer to layer.
Table~\ref{tab:unembed} reports the top-1 error rate when applying the unembedding layer on every layer using our test set from Section~\ref{sec:linear_reps}.
As a baseline, we apply the same unembedding layer from OthelloGPT to the residual streams of a randomly initialized GPT model. 
3) Linear probes across layers share similar directions.
Figure~\ref{fig:cos_sims} plots the cosine similarity between all linear probes, averaged across all 64 tiles and directions (\textsc{Mine}, \textsc{Yours}, \textsc{Empty}).


\begin{table*}
\center
\begin{tabular}{c|cccccccc}
\hline
  Baseline: Random &  $x^1$ & $x^2$ & $x^3$ & $x^4$ & $x^5$ & $x^6$ & $x^7$ & $x^8$ \\
\hline
0.856 & 0.215 & 0.152 & 0.112 & 0.079 & 0.049 & 0.015 & 0.004 & 0.001
\\
\hline
\end{tabular}
\caption{
Top-1 error rates when applying the unembedding layer to earlier layers.
As a baseline we apply OthelloGPT's unembedding layer on a randomly initialized GPT model.
}
\label{tab:unembed}
\end{table*}

\section{On Principled Ways of Probing}
\label{sec:appx_discussion_probing}

Probing has produced both excitement and skepticism amongst researchers \citep{belinkov-2022-probing}.
Here we provide our learnings regarding probing.

One criticism of probes is whether the discovered features are actually used by the model, i.e., correlation vs. causation.
Intervention is commonly used to study causality \citep{giulianelli-etal-2018-hood, tucker-etal-2021-modified}, but have often reached mixed conclusions \citep{belinkov-2022-probing}.
While both linear and non-linear probes have demonstrated successful interventions \citep{li2023inference, Turner_MacDiarmid_Udell_lisathiergart_Mini}, linear probes are much easier to interpret, as they imply that features simply correspond to vectoral directions.

Another challenge is knowing which features to probe for, which can lead to pitfalls.
Taking OthelloGPT as an example, classifying \{\textsc{Black}, \textsc{White}\} versus \{\textsc{Mine}, \textsc{Yours}\} leads to different takeaways, which illustrates the danger of \emph{projecting our preconceptions}.

Speaking of incorrect takeaways, our last point concerns the expressivity of probe models.
With an expressive-enough probe, there is a danger of the probe computing or memorizing the desired feature that one is looking for, rather than extracting \citep{pimentel-etal-2020-pareto, saphra-lopez-2019-understanding}.
Still, some researchers view linear classification as inadequate \citep{pimentel-etal-2020-information, saphra-lopez-2019-understanding}.
We view our work as evidence that linear probes do have interpretable and controllable power, and anticipate these findings to generalize to larger language models.


\end{document}